\documentclass{wscpaperproc}
\usepackage{latexsym}
\usepackage{graphicx}
\usepackage{mathptmx}
\usepackage[T1]{fontenc}

\usepackage{amsmath}
\usepackage{amsfonts}
\usepackage{amssymb}
\usepackage{amsbsy}
\usepackage{amsthm}

\usepackage[utf8]{inputenc}
\usepackage{tikz}
\usetikzlibrary{shapes, arrows.meta, positioning, calc}
\usepackage{enumitem}
\usepackage{xcolor}
\usepackage{colortbl}
\usepackage{collcell}
\usepackage{pgf}

\usepackage[pdftex,colorlinks=true,urlcolor=blue,citecolor=black,anchorcolor=black,linkcolor=black]{hyperref}

\usepackage[table]{xcolor}
\usepackage{pgfmath}
\usepackage{array}
\usepackage{tabularx} 
\usetikzlibrary{shadings}

\newcolumntype{H}{>{\centering\arraybackslash\scriptsize}p{1.4cm}}
\renewcommand{\arraystretch}{1.2}
\definecolor{myDarkRed}{RGB}{180, 20, 30} 
\definecolor{myLightBlue}{RGB}{180, 210, 255}

\newcommand{\HeatCell}[1]{%
  \pgfmathsetmacro{\val}{#1}%
  \ifdim\val pt < 50 pt
    \pgfmathsetmacro{\colormix}{(50-\val)*2}%
    \edef\x{\noexpand\cellcolor{blue!\colormix!white}}\x {\scriptsize #1}%
  \else
    \pgfmathsetmacro{\colormix}{(\val-50)*2}%
    \ifdim\val pt > 80 pt
        \edef\x{\noexpand\cellcolor{myDarkRed!\colormix!white}}\x \textcolor{white}{\scriptsize #1}%
    \else
        \edef\x{\noexpand\cellcolor{myDarkRed!\colormix!white}}\x {\scriptsize #1}%
    \fi
  \fi
}

\newtheoremstyle{wsc}
{3pt}
{3pt}
{}
{}
{\bf}
{}
{.5em}
{}

\theoremstyle{wsc}

\begin{document}

%
%

\pagestyle{fancyplain}

\thispagestyle{plain}
\firstPageHead{}

\chead{\fancyplain{}{\itshape Cinar, Ozkose, Von Hoene, Roess, Anderson, and Kavak}}

\rhead{}
\cfoot{}
\renewcommand{\headrulewidth}{0pt} 

\makeatletter
\let\@internalcite\cite
\def\cite{\def\@citeseppen{-1000}%
    \def\@cite##1##2{(##1\if@tempswa , ##2\fi)}%
    \def\citeauthoryear##1##2##3{##1 ##3}\@internalcite}
\def\citeNP{\def\@citeseppen{-1000}%
    \def\@cite##1##2{##1\if@tempswa , ##2\fi}%
    \def\citeauthoryear##1##2##3{##1 ##3}\@internalcite}
\def\citeN{\def\@citeseppen{-1000}%
    \def\@cite##1##2{##1\if@tempswa, ##2)\else{}\fi}%
    \def\citeauthoryear##1##2##3{##1 (##3)}\@citedata}
\def\citeA{\def\@citeseppen{-1000}%
    \def\@cite##1##2{(##1\if@tempswa , ##2\fi)}%
    \def\citeauthoryear##1##2##3{##1}\@internalcite}
\def\citeANP{\def\@citeseppen{-1000}%
    \def\@cite##1##2{##1\if@tempswa , ##2\fi}%
    \def\citeauthoryear##1##2##3{##1}\@internalcite}
\def\shortcite{\def\@citeseppen{-1000}%
    \def\@cite##1##2{(##1\if@tempswa , ##2\fi)}%
    \def\citeauthoryear##1##2##3{##2 ##3}\@internalcite}
\def\shortciteNP{\def\@citeseppen{-1000}%
    \def\@cite##1##2{##1\if@tempswa , ##2\fi}%
    \def\citeauthoryear##1##2##3{##2 ##3}\@internalcite}
\def\shortciteN{\def\@citeseppen{-1000}%
    \def\@cite##1##2{##1\if@tempswa, ##2\else{}\fi}%
    \def\citeauthoryear##1##2##3{##2 (##3)}\@citedata}
\def\shortciteA{\def\@citeseppen{-1000}%
    \def\@cite##1##2{(##1\if@tempswa , ##2\fi)}%
    \def\citeauthoryear##1##2##3{##2}\@internalcite}
\def\shortciteANP{\def\@citeseppen{-1000}%
    \def\@cite##1##2{##1\if@tempswa , ##2\fi}%
    \def\citeauthoryear##1##2##3{##2}\@internalcite}
\def\citeyear{\def\@citeseppen{-1000}%
    \def\@cite##1##2{(##1\if@tempswa , ##2\fi)}%
    \def\citeauthoryear##1##2##3{##3}\@citedata}
\def\citeyearNP{\def\@citeseppen{-1000}%
    \def\@cite##1##2{##1\if@tempswa , ##2\fi}%
    \def\citeauthoryear##1##2##3{##3}\@citedata}
%
%
%
\def\@citedata{%
    \@ifnextchar [{\@tempswatrue\@citedatax}%
                  {\@tempswafalse\@citedatax[]}%
}

\def\@citedatax[#1]#2{%
\if@filesw\immediate\write\@auxout{\string\citation{#2}}\fi%
  \def\@citea{}\@cite{\@for\@citeb:=#2\do%
    {\@citea\def\@citea{, }\@ifundefined
       {b@\@citeb}{{\bf ?}%
       \@warning{Citation `\@citeb' on page \thepage \space undefined}}%
{\csname b@\@citeb\endcsname}}}{#1}}%

%
\def\@citex[#1]#2{%
\if@filesw\immediate\write\@auxout{\string\citation{#2}}\fi%
  \def\@citea{}\@cite{\@for\@citeb:=#2\do%
    {\@citea\def\@citea{; }\@ifundefined
       {b@\@citeb}{{\bf ?}%
       \@warning{Citation `\@citeb' on page \thepage \space undefined}}%
{\csname b@\@citeb\endcsname}}}{#1}}%

%
\def\@biblabel#1{}
\makeatother



\newdimen\bibindent
\bibindent=0.0em
\def\thebibliography#1{\section*{\refname}\list
   {}{\settowidth\labelwidth{[#1]}
   \leftmargin\parindent
   \itemindent -\parindent
   \listparindent \itemindent
   \itemsep 0pt
   \parsep 0pt}
   \def\newblock{}
   \sloppy
   \sfcode`\.=1000\relax}


\setlength{\baselineskip}{12.7pt}

\title{Leveraging Large Language Models for Systematic Literature Review of Disease Spread Models}

\author{\begin{center}Orhan Yagizer Cinar\textsuperscript{1}, 
Timur Emre Ozkose\textsuperscript{2}, Emma Von Hoene\textsuperscript{3}, Amira Roess\textsuperscript{4}, Taylor Anderson\textsuperscript{3}, and Hamdi Kavak\textsuperscript{2}\\
[11pt]
\textsuperscript{1}Dept.~of Computer Science, George Mason University, Fairfax, VA, USA\\
\textsuperscript{2}Dept.~of Computational and Data Sciences, George Mason University, Fairfax, VA, USA\\
\textsuperscript{3} College of Public Health, George Mason University, Fairfax, VA, USA\\
\textsuperscript{4}Dept.~of Geography and Geoinformation Science, George Mason University, Fairfax, VA, USA\end{center}
}

\maketitle


\section*{ABSTRACT}
Recent advancements in Large Language Models (LLMs) have created new opportunities to streamline and potentially automate many research processes, including systematic literature reviews (SLRs). This study reports an LLM pipeline development for extracting model-relevant information from 536 peer-reviewed agent-based modeling papers. We compare the results with those of a human-conducted SLR. Our results show paper-level accuracies of approximately 77.95\% for GPT-4.1 and 81.67\% for GPT-5.0. Field-level accuracy ranges from 32.40\% to 100.00\%, with more complex or subjective fields performing less reliably. Importantly, we find that agreement between LLMs is a potential indicator of output quality: low agreement may signal hallucinations, whereas high agreement combined with low accuracy may point to noise or errors in the human dataset. Overall, our study provides practical insights into prompt development and highlights both the potential and limitations of using LLMs for full-scale SLRs in the modeling and simulation domain.

\section{Introduction}
\label{sec:intro}

Systematic literature reviews (SLR) identify and synthesize scientific evidence to answer a research question using transparent and reproducible methods \shortcite{lame2019systematic}. SLRs are widely used in many disciplines, including advancements in simulation models of disease spread \shortcite{nianogo2015agent,willem2017lessons}, socio-ecological systems \shortcite{nugroho2023systematic}, human mobility \shortcite{jing2020agent}, and policy modeling \shortcite{belfrage2024simulating}. SLRs follow rigorous protocols and reporting guidelines such as PRISMA \shortcite{page2021prisma}. Consequently, SLRs are research-intensive, both time-consuming and expensive. They require time for searching, screening, full-text assessment, and manual data extraction. Previous work quantified the labor effort in SLR, finding that a single SLR ranges from a minimum of 6-16 months, costing approximately \$141,194.80 \shortcite{michelson2019significant,borah2017analysis}. 

The rapid development of large language models (LLMs) has created new opportunities to reduce the time and costs to conduct SLRs. Accordingly, researchers have begun evaluating the performance of LLMs across multiple stages of the SLR workflow. A recent review identified 37 papers that investigate the use of LLMs for SLR.  LLM approaches covered most SLR steps, including literature search (n = 15, 41\%), study selection (n = 14, 38\%), and data extraction (n = 11, 30\%) \shortcite{lieberum2025large}. Validation studies that compare LLM results with human reference data made up about half of the studies (n = 21, 57\%). 

Within this broader application of LLMs to SLR, data extraction has attracted particular attention because it is one of the most labor-intensive steps and prone to reviewer subjectivity \shortcite{nussbaumer2021resource}. Existing studies that evaluate the potential integration of LLMs provide proof-of-concept results on small samples, typically for clinical SLRs \shortcite{mahmoudi2025critical,gartlehner2024data,konet2024performance,gartlehner2025artificial} (Section~\ref{sec:bg} provides more details). While the literature demonstrates the plausibility of LLM-based extraction for SLR, evidence remains limited on how well these approaches perform when scaled to the full volume of a systematic review, especially for modeling and simulation areas. For instance, the predominance of LLM applications to clinical SLR motivates the need to better understand LLM performance in other contexts, such as extracting simulation model characteristics (e.g., model purpose, calibration, uncertainty reporting). 

The objective of this study is to address this gap by evaluating LLM-based data extraction using GPT-4.1 and GPT-5.0 against a reference dataset from a previously completed SLR of 536 COVID-19 agent-based models (ABMs) \shortcite{von2026all}. Following the data schema of the reference dataset, we developed an automated LLM pipeline that processes full-text scientific papers and extracts structured outputs that can be compared to the human-extracted data on a field-by-field basis. 
This pipeline design allows for the comparison between human-extracted and LLM-extracted data while separating the extraction task from stages of the systematic review workflow, such as screening or synthesis.

The rest of this paper is structured as follows. Section~\ref{sec:bg} summarizes how Artificial Intelligence (AI) tools have been used to streamline and advance research. Section~\ref{sec:methods} presents the methods used in this study. What follows in Section~\ref{sec:results} is the comparison between human-extracted SLR data and LLM-extracted SLR data, and also a comparison across LLM outputs (GPT-5.0 vs. GPT-4.1). We conclude the paper in Section~\ref{sec:discussion} with the summary and discussion.

\section{Large Language Models in Systematic Literature Reviews} \label{sec:bg}

It has long been recognized that key components of the research enterprise can be formalized and supported computationally \shortcite{gil2014amplify}. Early work in the 1980s argued that important scientific discovery processes, such as hypothesis generation, model construction, and theory revision, follow known patterns, which make it possible to model them in a process~\shortcite{langley1987scientific}. Such ideas implicitly laid the groundwork to automate scientific activities, including literature review, knowledge synthesis, and new discoveries. As of this writing, a vast number of AI tools for accelerating science have been developed, some claim \shortcite{mitchener2025kosmos} to accomplish the vision that was put forth several decades ago. Given rapid advances in LLMs, it is now crucial to characterize the state of the art in using LLMs to answer SLR research questions.

According to the PRISMA (the Preferred Reporting Items for Systematic reviews and Meta-Analyses) guidelines, a typical SLR follows several required steps to be considered compliant. These steps include (1) defining the review rationale and objectives, (2) specifying eligibility criteria, (3) identifying information sources, (4) developing reproducible search strategies, (5) screening for study eligibility, (6) extracting data from included studies, (7) defining outcomes and other variables, (8) synthesizing, and reporting (9). LLMs have the potential to contribute to these steps in two fundamental ways. First, LLMs can assist the reviewers in streamlining different steps. For instance, when identifying eligibility criteria, LLMs can recommend relevant keywords and expand the original criteria to something more comprehensive. We can call this approach \textbf{``LLM-assisted''} process where humans are still in the loop and make decisions. Second, LLMs can completely replace humans. For instance, when deciding the eligibility of a study, LLMs have the ability to automatically determine if the study should be included based on given criteria. We can call this approach \textbf{``LLM-based''} process where humans are not directly included in the decision-making. A recent review by Lieberum et al.~\citeyear{lieberum2025large} reveals that LLMs have been used in the majority of the steps of SLRs in both LLM-assisted and LLM-based modalities, highlighting their potential.

Our focus in this work is on the data extraction from the included studies. Many LLM-based data extraction studies use widely accessible tools in a zero-shot setting, without task-specific fine-tuning or additional training data. For example, Mahmoudi et al.~\citeyear{mahmoudi2025critical} used the web-based ChatGPT prompts to extract data related to behavioral models in COVID-19 simulations from a sample of 10 studies, finding that its accuracy improves as prompts were refined, showing better performance in explicit data extraction (e.g., study setting) than subjective data. In a similar setting, Gartlehner et al.~\citeyear{gartlehner2024data} utilized the web-based Claude 2 to extract clinical study participant data from a sample of 10 studies, finding high overall accuracy (96\%). Additionally, their study reported high ease of use, requiring no technical expertise. Collectively, these studies indicate that a zero-shot commonly accessible LLM shows promising results.

Other studies extended data extraction tasks to multiple LLMs with different versions. Konet et al.~\citeyear{konet2024performance} used web-based Claude 2 and GPT-4 to extract clinical study participant data from 10 papers, finding accuracy was higher for Claude 2 while citing also technical challenges with GPT-4. Gartlehner et al.~\citeyear{gartlehner2025artificial}, on the other hand, utilized Claude's web-based interface (versions 2.1, 3.0 Opus, and 3.5 Sonnet) to extract data from 63 studies. They compare human data extraction with LLM assisted (human in the loop) data extraction, finding that LLM-assisted had a higher accuracy (91\% compared to 89\%). 

Overall, these studies show the potential of using LLMs in SLR data extraction, supporting the claims of Lieberum et al.~\citeyear{lieberum2025large}. However, the majority of the existing literature for SLR data extraction were tested on a small sample of articles utilizing less reproducible web-based LLM interfaces. Equally importantly, the focus on clinical studies limits the extension of these findings to other domains such as modeling and simulation. Our study is large-scale (N=536) with a focus on COVID-19 agent-based models, and has the potential to reveal insights that are crucial for the modeling and simulation community.   

\section{Methods}\label{sec:methods}

In this study, we assess the extent to which LLMs can replicate SLR data extraction at scale, using a reference human-extracted dataset from a systematic review of COVID-19 ABMs~\shortcite{von2026all}. We do this by developing an automated LLM pipeline processing each full-text paper and generating structured outputs aligned with the human extraction schema using GPT-4.1 and GPT-5.0 via the OpenAI Application Programming Interface (API). We quantify the similarity between LLM-generated and human-extracted data using the Jaccard index and an overlap coefficient, with results aggregated across extraction fields and across papers. The details for the human-extracted SLR data are described in Section~\ref{sec:refData} and the LLM pipeline for data extraction and evaluation are detailed in Section~\ref{sec:llm}.

\subsection{Human-Extracted Reference Data}\label{sec:refData}

This study compares SLR data extracted by LLMs with corresponding data extracted by human reviewers. The human reference dataset comes from a systematic review of COVID-19 agent-based modeling (ABM) papers (N=536) published between January 2020 and December 2023. Details of the original human-extracted SLR data are reported by Von Hoene et al.~\citeyear{von2026all}. The goal of the original study was to assess how well COVID-19 ABMs aligned with best practices for model usefulness, using criteria spanning model transparency and reuse, interdisciplinary collaboration and stakeholder engagement, and model evaluation. 

\begin{figure}[]
\centering
{\includegraphics[width=0.75\linewidth]{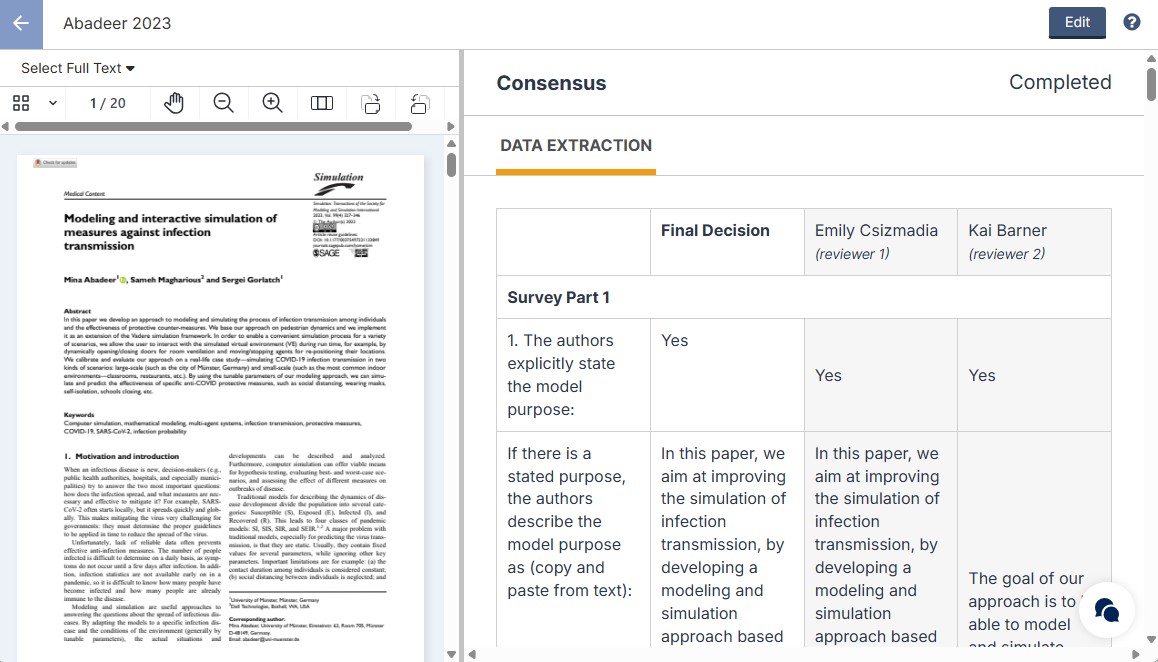}}
\caption{Covidence interface.}\label{fig:covidence}
\end{figure}

In brief, the authors developed a standardized extraction form to systematize data collection from each paper. Reviewers were trained on the extraction form during a pilot phase before full-scale extraction was conducted. Each paper was assigned to two independent reviewers, with assignments approximately balanced so that reviewers were paired with one another across the dataset. Covidence systematic review software was used to store source texts and extracted responses and to automatically flag discrepancies when paired reviewers entered conflicting values. The average human-human agreement was 74\%, ranging from 68.8\% (min) to 83.8\% (max). Consistent with SLR best practices, all disagreements were escalated to a third consensus reviewer (one of two lead researchers), who used Covidence’s side-by-side comparison interface to verify entries against the source text and adjudicate a final decision, as illustrated in Figure~\ref{fig:covidence}. The authors note that while Covidence supported workflow organization and conflict resolution, it did not automate any extraction judgments. With the exception of bibliographic metadata, all data were extracted manually and are therefore subject to reviewer interpretation.

The final dataset comprised 21 data fields with values determined through the consensus process (see Figure 3 for Questions). About half of the fields were binary (yes/no), identifying whether the authors: 1) explicitly stated the model purpose; 2) state that the model was or could be used by others (e.g., policy makers, decision makers); 3) specified the model to a study region; 4) used an open-source ABM platform and package (e.g., Repast, NetLogo, Mesa); 5) documented the model using standardized protocols (e.g., ODD, ODD+D, TRACE); 6) clearly stated assumptions and limitations; 7) made code available for download; 8) extended or reused an existing model; 9) involved multiple disciplines; 10) incorporated stakeholder feedback in the modeling cycle; 11) reported performing model evaluation (verification, validation, sensitivity testing, validation); and 12) quantified and communicated the uncertainty. Corresponding to each yes/no binary field, reviewers also copied-and-pasted semi-structured supporting evidence from the text (e.g., name of study region, ABM platform, reporting protocol). Overall, the full-scale extraction of 536 papers took approximately one year (February-December 2024), underscoring the high cost of such large-scale manual data extraction, even with the support of systematic review software.

\subsection{LLM Pipeline}\label{sec:llm}

\begin{figure}[htbp]
\centering
\resizebox{\textwidth}{!}{%
    \begin{tikzpicture}[
        node distance = 2cm and 0.8cm, 
        process/.style = {
            rectangle,
            draw=black!60,
            thick,
            text width=5.4cm,
            align=left,
            font=\small,
            inner sep=0.3cm,
            anchor=center
        },
        arrow/.style = {
            -{Stealth[length=3mm]},
            thick,
            draw=gray!50
        }
    ]
    \newcommand{\boxcontent}[2]{%
        \begin{minipage}[t][4.0cm]{5.5cm}
            \textbf{#1}
            \begin{itemize}[leftmargin=0.3cm, labelsep=0.1cm, nosep, topsep=2pt, partopsep=0pt, label=\textbullet]
                #2
            \end{itemize}
        \end{minipage}%
    }

    \node (prep) [process] {
        \boxcontent{Preparation}{
            \item Human-extracted reference dataset from a previously completed systematic review of 536 COVID-19 agent-based models.
            \item Fixed categorical extraction schema defined based on human-reviewed data.
            \item Schema used as the reference standard for LLM-based extraction.
        }
    };

    \node (testing) [process, right=of prep] {
        \boxcontent{Prompt Design}{
            \item Human extraction form translated into a structured LLM-compatible prompt.
            \item Fixed response options specified for all variables.
            \item Strict requirement for JSON only output with one key per field.
        }
    };

    \node (reliability) [process, right=of testing] {
        \boxcontent{LLM-Based Extraction}{
            \item Full-text scientific papers provided as PDF inputs.
            \item One paper processed per API request using GPT-4.1 and GPT-5.0 models.
            \item Batch processing applied to manage rate limits and ensure reproducibility.
        }
    };

    \node (screening) [process, below=of prep] {
        \boxcontent{Output Validation}{
            \item Automated validation of model outputs against the required JSON schema.
            \item Rule-based repair applied to malformed outputs.
            \item Secondary LLM-based recovery used when rule-based repair failed.
        }
    };

    \node (extraction) [process, right=of screening] {
        \boxcontent{LLM–Human Comparison}{
            \item Examined LLM outputs aligned with corresponding human records at the paper level.
            \item Field-level Jaccard and Coverage similarity computed for categorical and multi-valued variables.
            \item Comparison results prepared for downstream quantitative analysis.
        }
    };

    \node (analysis) [process, right=of extraction] {
        \boxcontent{Statistical Analysis}{
            \item Field-level scores aggregated to paper-level accuracy measures.
            \item Descriptive accuracy statistics computed across all papers.
            \item Statistical significance assessed using non-parametric tests.
        }
    };

    \draw [arrow] (prep) -- (testing);
    \draw [arrow] (testing) -- (reliability);
    \draw [arrow] (screening) -- (extraction);
    \draw [arrow] (extraction) -- (analysis);

    \draw [arrow] (reliability.south) -- ++(0,-1) -| (screening.north);

    \end{tikzpicture}%
    }
\caption{Overview of the LLM-based data extraction and evaluation pipeline.}
\label{fig:ai-extraction-workflow}
\end{figure}
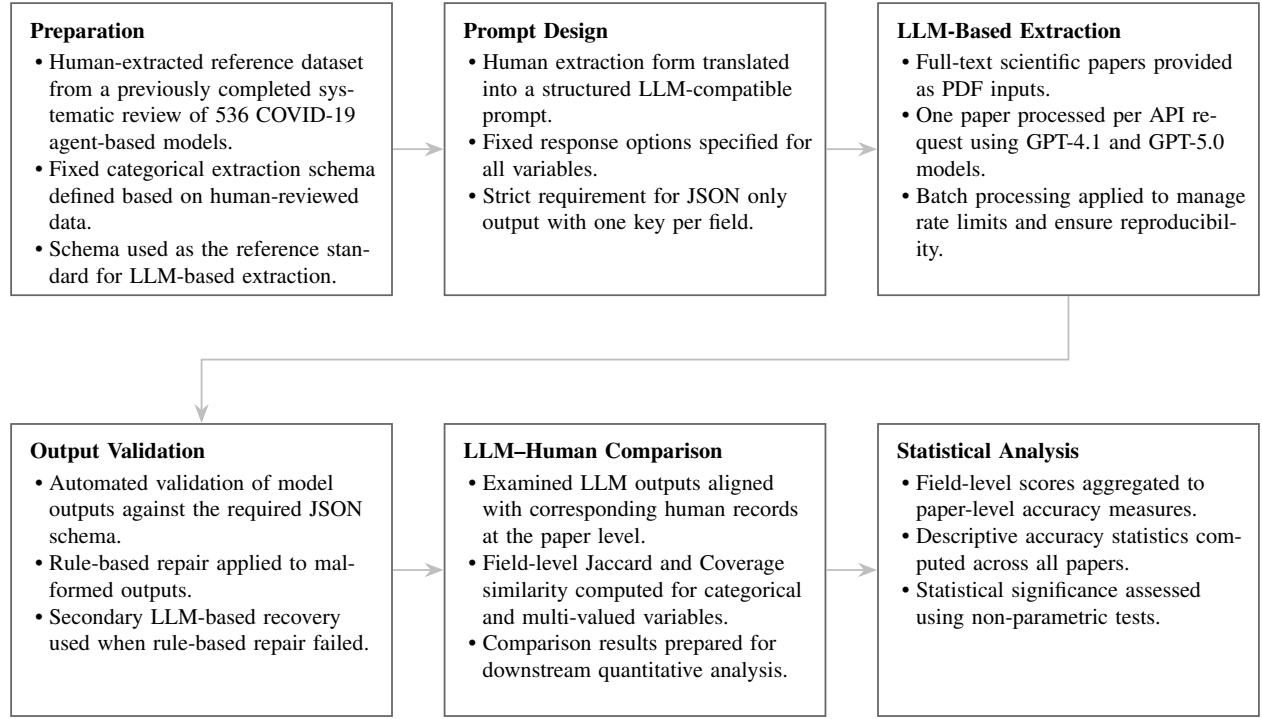

We implemented a data extraction pipeline using the OpenAI API \shortcite{OpenAI2026}, enabling fully programmatic and reproducible interaction with LLMs. The overall pipeline is summarized in Figure~\ref{fig:ai-extraction-workflow}, which shows six main steps. All experiments were conducted via secure API calls, unlike prior work that relied on web interfaces (see Section~\ref{sec:bg}), to ensure reliable execution, controlled batching, and detailed logging of model usage. This approach allowed each paper to be processed independently and enabled scaling the extraction pipeline across the full dataset.

We used GPT-4.1 and GPT-5.0, which are stable LLMs that were widely accessible to the research community at the time of this study. Both models were used in a zero-shot configuration, without any adaptation or tuning on the human-extracted dataset. This was a deliberate decision to understand the capabilities and limitations of LLMs under default usage conditions. The model parameters were set to API defaults, with temperature and top-p set to 1.0, and frequency and presence penalties set to 0.0. While these are not minimum stochasticity settings, the constrained JSON schema with fixed categorical response options substantially limits output variance in practice. We acknowledge the absence of replicate runs as a limitation and recommend that future work include runs on a sample to formally quantify the sampling variance.

A single structured prompt was applied uniformly to all papers. The prompt\footnote{See the notes section in the end to access the pipeline code.}, which was designed to replicate the original form for data extraction, included: (i) detailed instructions describing each data extraction variable, (ii) specific permissible categorical response choices for each field, and (iii) specific restrictions to ensure that only one valid JavaScript Object Notation (JSON) object with a single key for each data extraction variable is returned. The prompt specifically excluded any text for explanation, commentary, or generation of new categories outside those specified in the schema.

Full-text scientific papers were provided to the models in the form of uploaded PDF files, with each paper being processed by one API call. To deal with the processing rate, papers were processed in batches with time delays between API calls. One API call provided one structured extraction result for one paper, providing a direct mapping between LLM-extracted results and human-extracted results.

\vspace{5mm}

Because evaluation required strictly structured output, the validation of each model response was done automatically to ensure compatibility with a specified JSON schema, using a three-step process. In the first step, the model output was parsed directly as JSON. If parsing failed, a second step applied rule-based corrections targeting only structural formatting errors, specifically missing or mismatched brackets, quotation marks, and trailing commas. If rule-based repair also failed, a third step invoked a lightweight LLM (GPT-4.1-mini) with an explicit instruction to correct only the JSON formatting. Critically, no extracted content values were modified at any stage; all repairs were limited to JSON syntax. All 536 papers produced valid structured output after this pipeline. 
The resulting outputs from all extractions were stored in JSON format and converted for comparison with the human extracted data. Use of tokens and request parameters was tracked to enable reproducibility. Across both models, 1,072 API calls were made (one call per paper per model). Based on tracked token usage, the pipeline ran at well under \$10 per model run, making this approach highly cost-effective relative to the cost of human SLRs, estimated at tens of thousands of dollars. 

Both GPT models support context windows that exceed the length of any individual paper in the corpus. GPT-4.1 supports up to 1,047,576 tokens, while GPT-5.0 supports up to 400,000 tokens (OpenAI, 2026). Based on token usage tracked per API call, no paper required truncation; each was processed in full within a single request. Full-text PDFs were provided directly via the OpenAI File Upload API, which handles PDF parsing internally on the backend before passing document content to the model. 
While this approach is convenient and eliminates preprocessing, it also introduces a limitation: API's internal PDF parser may not completely capture all document elements, particularly multi-column layouts, embedded tables, footnotes, and supplementary appendices. Extraction errors in fields such as author disciplines, ABM platform names, or model code availability may therefore partly reflect PDF parsing artifacts rather than model reasoning failures. Our results may improve in the future depending on the changes in the PDF parser.

\subsection{Evaluation Logic}

\begin{figure*}[h]
\centering
\includegraphics[
  width=\textwidth
]{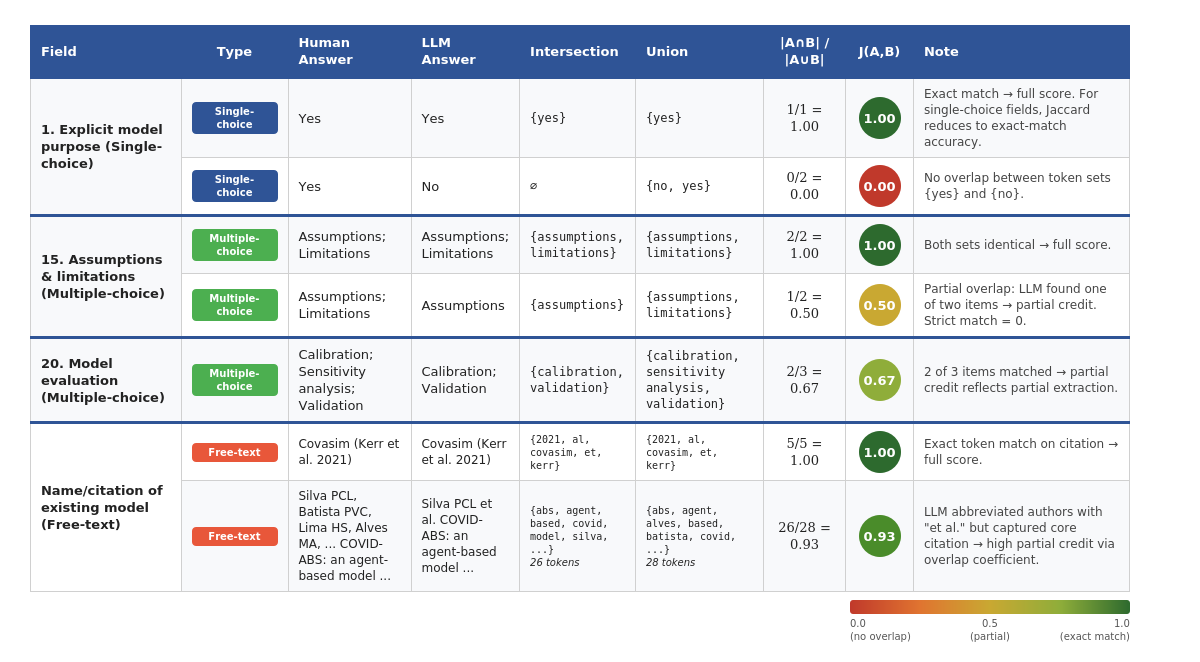}
\caption{Examples of accuracy score calculations for different field types. }
\label{fig:jaccard}
\end{figure*}

Prior to comparing the LLM against the human reference data, all the text was converted to
lowercase and punctuation was removed. There are three types of fields, including single-choice (only one word/token), multi-choice (check all that apply), and free-text. 
The presence of multi-word phrases in multi-select and free-text fields, such as ``United States'', creates an assessment challenge. Exact string match methods do not recognize conceptually equivalent data based on wording and ordering. To address this, we applied field-type-specific similarity metrics, which are exemplified in Figure~\ref{fig:jaccard}.

\begin{itemize}
    \item For \textbf{multi-select fields} (e.g., disciplines, assumptions \& limitations, or evaluation methods),
the Jaccard index was used to evaluate the degree of overlap between sets:
\begin{equation}
J(A,B)=\frac{|A\cap B|}{|A\cup B|}
\end{equation}
where $A$ and $B$ are the token sets derived from the two compared responses. The
score lies in the interval $[0,1]$, where 1 implies identical sets and 0 implies no
overlap. To ensure that multi-word expressions (e.g., ``United Kingdom'' or
``Sensitivity analysis'') are treated as single conceptual units, the text was
tokenized using structural delimiters (semicolons, commas, pipes) as opposed to
whitespace.\\

\item For \textbf{free-text fields} (e.g., country, model name, name of the platform, or the name of the existing model), the overlap coefficient was used instead:
\begin{equation}
\text{Overlap}(A,B)=\frac{|A\cap B|}{\min(|A|,|B|)}
\end{equation}

This metric is more forgiving than Jaccard when one response is a subset of the
other (e.g., an LLM omitting minor tokens from a citation), providing proportionate
partial credit rather than penalizing length differences.

\end{itemize}

As free-text and multi-select fields are evaluated using different similarity metrics, their scores are not directly comparable across field types in Table~\ref{tab:accuracy} while they operate within the same numerical range. For single-choice variables (i.e., choose only one from multiple options), both metrics collapse into a typical accuracy measure. Because sets $A$ and $B$ each contain a single element after preprocessing, a correctly extracted response yields a score of 1.0 (100\% accuracy) and a mismatched response yields 0.0 (0\% accuracy). Additionally, when both the human reference and LLM output are empty for a given field, the score is 1.0, reflecting correct agreement on the absence of information. Details about all field types and their evaluation are shared as a reference document in the online repository.

\section{Results} \label{sec:results}
This section reports our findings, highlighting the extent to which LLM-extracted data match human-extracted data and the conditions under which the models fail to do so. We divide the analysis into two parts based on the accuracy at (1) the field level and (2) the individual-paper level. The following sections address these in turn.

\subsection{Field-level Accuracy}
For each field, we calculated summary statistics across all 536 papers. The heatmap in Table~\ref{tab:heatmap} reports the accuracy for each field between each LLM and the human extracted reference data, as well as the agreement between GPT-4.1 and GPT-5.0 models. The field ``The authors explicitly state the model purpose'' had the highest accuracy score in each comparison (100\% for each model comparison), while the ``Sub-classifications'' field for author disciplines had the lowest accuracy score with 32.4\% for GPT-4.1, 37.4\% for GPT-5.0, and 49.4\% agreement between GPT-4.1 vs.\ GPT-5.0. Note that all 536 papers in the human reference dataset were labeled 'Yes' on this field, resulting in zero class variance; while the LLM independently reached the same conclusion in zero-shot mode, chance-corrected statistics such as Cohen's kappa cannot be computed here, and this field should not be interpreted as a discriminative test of extraction capability. In a similar vein, low accuracy (<60\%) and limited agreement between models (67.1\%) was observed in the field ``Main disciplines.'' For the other fields, the accuracy and agreement scores were mostly above 65\%. 

For fields with high scores, we can conclude that LLMs can be impressively accurate and useful to understand model relevant information. This includes the statement of model purpose, whether the model is specific to a geographic region, use of any standard documenting protocol, and the statement of model assumptions and limitations. At the very least, this can accelerate the SLR process without sacrificing too much from accuracy.

When it comes to performances of different field types, results show that the LLMs tend to have higher accuracy and agreement for free-text fields with 81.6\% for GPT-4.1 vs Human, 85.0\% for GPT-5.0 vs Human. Single-choice fields come second with accuracies of 81.2\% for GPT-4.1 vs Human, 84.0\% for GPT-5.0 vs Human. For the multi-choice fields, the accuracy and agreement were lower with 58.3\% for GPT-4.1 vs Human, 66.0\% for GPT-5.0 vs Human. This reflects the complexity of "check all that apply" tasks, particularly for subjective fields like sub-classifications and model evaluation.\\

\begin{table}[ht]
\caption{Heatmap of accuracy across fields for Human vs Human and LLM Models and the agreement
between GPT-4.1 and GPT-5.0. \textbf{\textit{(SC)}} indicates Single-Choice field; \textbf{\textit{(MC)}} indicates Multiple-Choice field; \textbf{\textit{(FT)}} indicates Free-Text field.}
\label{tab:accuracy}
\centering
\small 
\renewcommand{\arraystretch}{1.2}
\setlength{\tabcolsep}{0pt} 
\setlength{\arrayrulewidth}{0pt} 
\arrayrulecolor{white}

\begin{tabular}{r | H | H | H |}
\multicolumn{1}{c}{} & 
\multicolumn{1}{c}{\rotatebox{80}{GPT-4.1 vs Human}} & 
\multicolumn{1}{c}{\rotatebox{80}{GPT-5.0 vs Human}} & 
\multicolumn{1}{c}{\rotatebox{80}{GPT-4.1 vs GPT-5.0}} \\ \cline{2-4}

The authors explicitly state the model purpose \textbf{\textit{(SC)}}:~ & \HeatCell{100.0} & \HeatCell{100.0} & \HeatCell{100.0} \\ \cline{2-4}
Main purpose of the article \textbf{\textit{(SC)}}:~ & \HeatCell{72.6} & \HeatCell{70.0} & \HeatCell{76.0} \\ \cline{2-4}
Sub-purpose of the article \textbf{\textit{(SC)}}:~ & \HeatCell{63.0} & \HeatCell{61.8} & \HeatCell{67.5} \\ \cline{2-4}
The authors state that the model can be/will be/has been used by others \textbf{\textit{(SC)}}:~ & \HeatCell{72.8} & \HeatCell{80.4} & \HeatCell{78.9} \\ \cline{2-4}
The model is specific to a study region \textbf{\textit{(SC)}}:~ & \HeatCell{89.4} & \HeatCell{93.5} & \HeatCell{86.2} \\ \cline{2-4}
If the model is specific to a study region, the study area country is \textbf{\textit{(FT)}}:~ & \HeatCell{81.0} & \HeatCell{85.8} & \HeatCell{83.1} \\ \cline{2-4}
The model is developed using open-source ABM platforms \textbf{\textit{(SC)}}:~ & \HeatCell{76.1} & \HeatCell{86.8} & \HeatCell{87.9} \\ \cline{2-4}
Name the platform or package \textbf{\textit{(FT)}}:~ & \HeatCell{72.3} & \HeatCell{79.9} & \HeatCell{85.1} \\ \cline{2-4}
The model is documented using standard protocols \textbf{\textit{(SC)}}:~ & \HeatCell{98.3} & \HeatCell{98.0} & \HeatCell{99.3} \\ \cline{2-4}
Name the protocol used \textbf{\textit{(FT)}}:~ & \HeatCell{98.3} & \HeatCell{98.0} & \HeatCell{99.3} \\ \cline{2-4}
The model assumptions and limitations are clearly stated \textbf{\textit{(MC)}}:~ & \HeatCell{94.3} & \HeatCell{94.8} & \HeatCell{98.5} \\ \cline{2-4}
The model code is available for download \textbf{\textit{(SC)}}:~ & \HeatCell{79.1} & \HeatCell{87.1} & \HeatCell{89.4} \\ \cline{2-4}
Provide the link where model is available \textbf{\textit{(FT)}}:~ & \HeatCell{87.4} & \HeatCell{89.7} & \HeatCell{92.0} \\ \cline{2-4}
The model extends or uses an existing model \textbf{\textit{(SC)}}:~ & \HeatCell{77.6} & \HeatCell{80.0} & \HeatCell{87.1} \\ \cline{2-4}
Name/citation of the existing model used \textbf{\textit{(FT)}}:~ & \HeatCell{68.8} & \HeatCell{71.8} & \HeatCell{77.6} \\ \cline{2-4}
How many disciplines \textbf{\textit{(SC)}}:~ & \HeatCell{77.9} & \HeatCell{75.8} & \HeatCell{71.9} \\ \cline{2-4}
Main disciplines \textbf{\textit{(MC)}}:~ & \HeatCell{52.5} & \HeatCell{60.2} & \HeatCell{67.1} \\ \cline{2-4}
Sub-classifications \textbf{\textit{(MC)}}:~ & \HeatCell{32.4} & \HeatCell{37.4} & \HeatCell{49.4} \\ \cline{2-4}
The study design incorporates stakeholder feedback \textbf{\textit{(SC)}}:~ & \HeatCell{87.9} & \HeatCell{91.4} & \HeatCell{90.1} \\ \cline{2-4}
The authors state that they perform model evaluation \textbf{\textit{(MC)}}:~ & \HeatCell{53.9} & \HeatCell{71.4} & \HeatCell{62.3} \\ \cline{2-4}
Uncertainty in model predictions is quantified and communicated \textbf{\textit{(SC)}}:~ & \HeatCell{79.3} & \HeatCell{83.0} & \HeatCell{82.5} \\ \cline{2-4}
\end{tabular}

\vspace{1.5em} 
\hfill 
\begin{tikzpicture}[baseline=(current bounding box.center)]
    \shade[left color=blue, right color=white] (0,0) rectangle (2, 0.4);
    \shade[left color=white, right color=myDarkRed] (2,0) rectangle (4, 0.4);
    
    \draw (0,0) rectangle (4, 0.4);
    
    \foreach \x/\label in {0/0, 1/25, 2/50, 3/75, 4/100} {
        \draw (\x, 0) -- (\x, -0.1) node[below, font=\tiny] {\label};
    }
    
    \node[anchor=west, font=\scriptsize] at (4.2, 0.2) {Accuracy Score (\%)};
\end{tikzpicture}

\label{tab:heatmap}

\end{table}



\subsection{Paper-level Accuracy}
\begin{figure*}[h]
\centering
\includegraphics[
  width=\textwidth
]{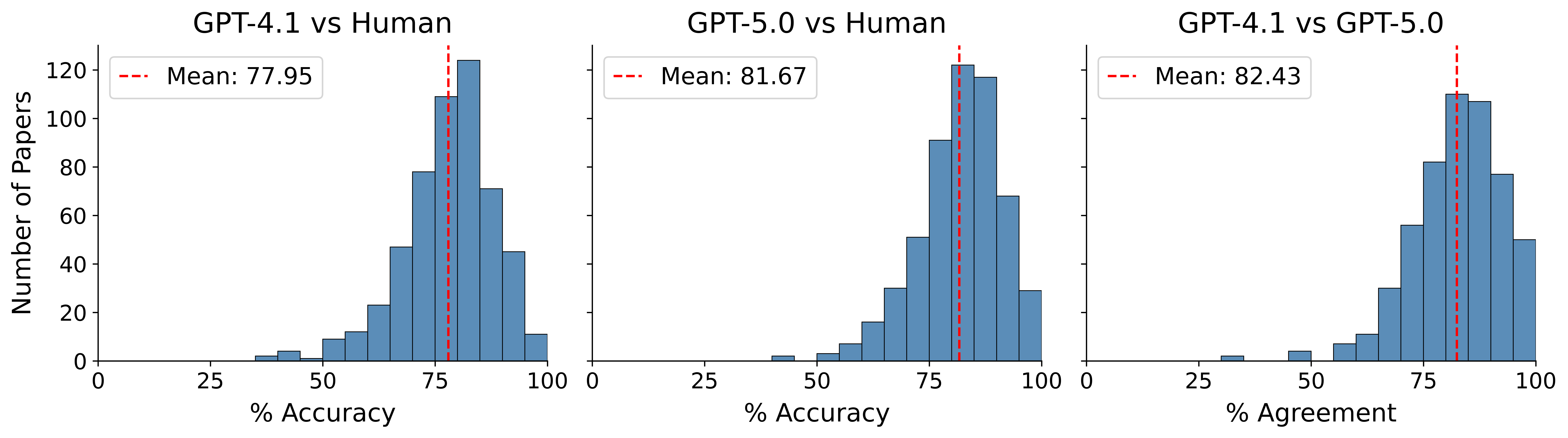}
\caption{Distribution of accuracy across fields for Human vs GPT-4.1 and Human vs GPT-5.0; and the agreement between GPT-4.1 and GPT-5.0.}
\label{fig:histogram}
\end{figure*}

We calculated \textit{paper-level accuracy} scores by averaging the field-level scores between the LLM and the human reference for each paper. We visualize the distribution of these paper-level accuracy scores in Figure~\ref{fig:histogram}. The first panel (left) indicates that a larger proportion of papers extracted by GPT-4.1 exhibit lower accuracy relative to the upper range of values. In contrast, the distribution of the second panel (middle) is shifted toward higher accuracy for GPT-5.0, with a greater concentration of papers achieving accuracy levels in the 80-90\% range. The mean accuracies are 77.95\% ($SD$ = 10.24\%) for GPT-4.1 vs.\ Human, 81.67\% ($SD$ = 9.28\%) for GPT-5.0 vs.\ Human, and 82.43\% ($SD$ = 10.01\%) for the agreement between GPT-4.1 and GPT-5.0.

To decide if the performance difference between the models was statistically significant, we conducted paired comparisons of per-paper accuracy for GPT-4.1 and GPT-5.0 across all 536 papers. Both GPT-4.1 and GPT-5.0 were evaluated against the same human reference dataset, yielding 536 naturally paired observations with no data exclusions. A paired samples $t$-test revealed a statistically significant difference in extraction accuracy
($t(535) = 6.44$, $p < .001$, Cohen's $d = 0.28$), with GPT-5.0 outperforming GPT-4.1 by a mean difference of $+3.73$ percentage points ($SD$ = 13.39\%). This test demonstrates that although both models show high overall agreement (82.43\%), they do not always produce identical outputs for the same papers, suggesting differences in extraction behavior. Specifically, $d = 0.28$ indicates a small to medium effect size while high statistical significance ($p < .001$) is established. In other words, GPT-5.0's performance is greater than GPT-4.1 but not dramatically better.

\section{Discussion, Conclusion, and Future Work} \label{sec:discussion}

This study evaluates the extent to which state-of-the-art LLMs can replicate SLR data extraction at scale using a reference dataset from a review of 536 COVID-19 agent-based models (ABMs). Overall, our results show that LLMs can extract a substantial portion of structured SLR data with moderate-to-high agreement relative to a human reference dataset, achieving mean paper-level accuracies of 77.95\% for GPT-4.1 and 81.67\% for GPT-5.0. A paired samples $t$-test confirmed that GPT-5.0's improvement over GPT-4.1 was statistically significant ($t(535) = 6.44$, $p < .001$). These findings provide evidence that zero-shot LLM-based data extraction can be feasible even when applied to the full volume of a large-scale SLR study. Previous studies were conducted on small samples and faced reproducibility challenges due to reliance on web interfaces, whereas our study is significantly larger in scale and is reproducible via APIs.

Field-level performance patterns suggest that LLM extraction quality is strongly dependent on the structure and subjectivity of the extraction variable. For example, single-choice fields achieved higher accuracy and agreement than multi-choice fields. This aligns with the intuition that single-choice decisions often map to explicit statements in the paper, while multi-choice classifications require interpretation, and sometimes domain-specific judgment. In this data extraction schema, the multi-choice fields required a more nuanced interpretation or classification, such as ``Assumptions and Limitations'', ``Main disciplines'', and ``Sub-classifications''. These fields likely combine multiple sources of difficulty, including ambiguity in author roles, implicit disciplinary backgrounds, and inconsistent terminology across papers. Such variables may be inherently more difficult to operationalize even for human reviewers, suggesting that lower LLM performance may reflect not only model limitations but also limitations in the clarity and standardization of the underlying extraction schema. Surprisingly, free-text fields achieved similar average accuracy and agreement to single-choice fields. However, this is likely due to the straightforward nature of the data that was extracted in this way (e.g., study area country, name of platform or package, link where the model is available). 

A notable contribution of this work is the use of inter-model (GPT-4.1 vs. GPT-5.0) agreement as an additional dimension for evaluating the results. Fields where LLM-LLM agreement exceeds LLM-human agreement may point to issues in the human reference dataset (e.g., inconsistencies or contamination), since the model is internally consistent but diverges from the reference. For instance, in our updated results, the field "The model extends or uses an existing model" shows higher inter-model agreement (87.1\%) than either model's agreement with the human reference (77.6\% and 80.0\%), consistent with this interpretation. We note, however, that no field in this corpus exhibits an extreme low-human/high-model divergence, so we present inter-model agreement as a proposed diagnostic rather than a demonstrated correction of the reference. In contrast, fields where LLM-LLM agreement is low may indicate instability, sensitivity to prompt interpretation, or hallucination in the model outputs. This can be seen with the 'Sub-classifications' of the author disciplines. Together, these patterns suggest that it can be valuable to report LLM-LLM agreement alongside human-LLM agreement when evaluating extraction quality. This is particularly relevant for large-scale SLR settings, where the human reference dataset is itself the product of a complex workflow with multiple reviewers. Furthermore, evaluation frameworks that treat human labels as perfectly correct may overstate LLM errors.



Finally, it is important to interpret these findings in light of the deliberate research design made in this study. We used a zero-shot configuration without fine-tuning or adaptation and set conservative parameters to reduce stochasticity. This design supports reproducibility and provides a baseline estimate of what is achievable under default usage conditions. However, it also means that our reported performance likely underestimates what could be achieved with domain-adapted prompting strategies, few-shot examples, retrieval-augmented extraction, or LLM-assisted workflows.

Beyond scale, this study makes two contributions that extend prior work in ways that go beyond domain shift. First, applying zero-shot LLM extraction to a modeling and simulation SLR rather than the clinical settings (which dominate the existing literature), demonstrates that accuracy is not unique to clinical data, where fields tend to be more standardized and terminology more consistent. Second, the use of inter-model agreement as an independent evaluation potentially reframes what LLM-based extraction can do. When two models agree with each other but diverge from the human reference, the more consistent explanation is not necessarily model error; it may reflect inconsistency in the human reference itself. This is a dimension of SLR quality that traditional validation frameworks cannot detect, and it suggests that LLMs may be useful not only for replicating human extraction but for auditing it. We also do not rule out the possibility that such agreement might be due to training data overlap, a prompt artifact, or a PDF parser artifact.

Some of the 536 papers may have appeared in the pretraining corpus of GPT-4.1 or GPT-5.0, which could inflate accuracy on fields where memorized content drives the output. This cannot be verified with current API access, as OpenAI does not disclose the training data at the document level. However, two observations suggest it is not a primary driver. First, the human reference was produced through trained reviewer interpretation, so LLM outputs are compared against judgment-based labels rather than verbatim text. Second, the least-performing fields are precisely where memorization would be least useful, as they require contextual judgment rather than recall. We acknowledge this as a limitation and recommend that future work incorporate contamination detection approaches where feasible.

Future work can extend this study in several directions. First, we plan to further investigate \textit{inter-model agreement} as an evaluation signal by utilizing ensembles of different LLMs. This will help quantify stability, identify systematically ambiguous fields, and flag papers that may require human adjudication. Second, future work should explore \textit{fine-tuning and adaptation strategies}, including supervised fine-tuning on subsets of human-extracted examples, prompt tuning, and few-shot prompting, to improve performance on fields with low scores. Finally, we aim to expand beyond data extraction and cover a wider spectrum of established systematic review processes aligned with PRISMA, including initial screening and study selection, synthesis, and reporting, enabling a more comprehensive assessment of how LLMs can contribute to end-to-end systematic review workflows.

\section*{ACKNOWLEDGMENTS}
This research was funded by National Science Foundation’s Division of Environmental Biology (Award No. 2109647) and Division of Information and Intelligent Systems (Award No. 2302970).

\section*{Notes}\label{sec:notes}
\begin{itemize}
    \item We thank Sara Von Hoene, Szandra Peter, Ethan Hopson, Emily Csizmadia, Faith Fenyk, and Kai Barner, who worked on extracting the original data.
    \item Study titles, the Python code used in the pipeline development, and detailed scoring logic reference can be viewed and downloaded at \url{https://osf.io/mtrcp} (DOI: 10.17605/OSF.IO/MTRCP).
\end{itemize}

\footnotesize

\bibliographystyle{wsc}

\bibliography{assets/demobib}

@article{von2026all,
  title={All Models Are Wrong, but Can They Be Useful? Lessons from COVID-19 Agent-Based Models: A Systematic Review},
  author={Von Hoene, Emma and Von Hoene, Sara and P{\'e}ter, Szandra and Hopson, Ethan and Csizmadia, Emily and Fenyk, Faith and Barner, Kai and Leslie, Timothy and Kavak, Hamdi and Z{\"u}fle, Andreas and others},
  journal={Journal of Artificial Societies and Social Simulation},
  volume={29},
  number={1},
  year={2026},
  publisher={JASSS}
}

@article{mahmoudi2025critical,
  title={Critical Assessment of Large Language Models’(ChatGPT) Performance in Data Extraction for Systematic Reviews: Exploratory Study},
  author={Mahmoudi, Hesam and Chang, Doris and Lee, Hannah and Ghaffarzadegan, Navid and Jalali, Mohammad S},
  journal={JMIR AI},
  volume={4},
  number={1},
  pages={e68097},
  year={2025},
  publisher={JMIR Publications Inc., Toronto, Canada}
}

@article{gartlehner2024data,
  title={Data extraction for evidence synthesis using a large language model: A proof-of-concept study},
  author={Gartlehner, Gerald and Kahwati, Leila and Hilscher, Rainer and Thomas, Ian and Kugley, Shannon and Crotty, Karen and Viswanathan, Meera and Nussbaumer-Streit, Barbara and Booth, Graham and Erskine, Nathaniel and others},
  journal={Research synthesis methods},
  volume={15},
  number={4},
  pages={576--589},
  year={2024},
  publisher={Wiley Online Library}
}

@article{lieberum2025large,
  title={Large language models for conducting systematic reviews: on the rise, but not yet ready for use—a scoping review},
  author={Lieberum, Judith-Lisa and Toews, Markus and Metzendorf, Maria-Inti and Heilmeyer, Felix and Siemens, Waldemar and Haverkamp, Christian and B{\"o}hringer, Daniel and Meerpohl, Joerg J and Eisele-Metzger, Angelika},
  journal={Journal of Clinical Epidemiology},
  volume={181},
  pages={111746},
  year={2025},
  publisher={Elsevier}
}

@article{konet2024performance,
  title={Performance of two large language models for data extraction in evidence synthesis},
  author={Konet, Amanda and Thomas, Ian and Gartlehner, Gerald and Kahwati, Leila and Hilscher, Rainer and Kugley, Shannon and Crotty, Karen and Viswanathan, Meera and Chew, Robert},
  journal={Research synthesis methods},
  volume={15},
  number={5},
  pages={818--824},
  year={2024},
  publisher={Wiley Online Library}
}

@article{gartlehner2025artificial,
  title={Artificial Intelligence--Assisted Data Extraction With a Large Language Model: A Study Within Reviews},
  author={Gartlehner, Gerald and Kugley, Shannon and Crotty, Karen and Viswanathan, Meera and Dobrescu, Andreea and Nussbaumer-Streit, Barbara and Booth, Graham and Treadwell, Jonathan R and Han, Jung Min and Wagner, Jesse and others},
  journal={Annals of Internal Medicine},
  volume={178},
  number={12},
  pages={1763--1771},
  year={2025},
  publisher={American College of Physicians}
}

@article{borah2017analysis,
  title={Analysis of the time and workers needed to conduct systematic reviews of medical interventions using data from the PROSPERO registry},
  author={Borah, Rohit and Brown, Andrew W and Capers, Patrice L and Kaiser, Kathryn A},
  journal={BMJ open},
  volume={7},
  number={2},
  pages={e012545},
  year={2017},
  publisher={British Medical Journal Publishing Group}
}

@article{michelson2019significant,
  title={The significant cost of systematic reviews and meta-analyses: a call for greater involvement of machine learning to assess the promise of clinical trials},
  author={Michelson, Matthew and Reuter, Katja},
  journal={Contemporary clinical trials communications},
  volume={16},
  pages={100443},
  year={2019},
  publisher={Elsevier}
}

@article{nussbaumer2021resource,
  title={Resource use during systematic review production varies widely: a scoping review},
  author={Nussbaumer-Streit, Barbara and Ellen, Moriah and Klerings, Irma and Sfetcu, Raluca and Riva, Nicoletta and Mahmi{\'c}-Kaknjo, Mersiha and Poulentzas, Georgios and Martinez, P and Baladia, Eduard and Ziganshina, Liliya Eugenevna and others},
  journal={Journal of clinical epidemiology},
  volume={139},
  pages={287--296},
  year={2021},
  publisher={Elsevier}
}

@article{gil2014amplify,
  title={Amplify scientific discovery with artificial intelligence},
  author={Gil, Yolanda and Greaves, Mark and Hendler, James and Hirsh, Haym},
  journal={Science},
  volume={346},
  number={6206},
  pages={171--172},
  year={2014},
  publisher={American Association for the Advancement of Science}
}

@book{langley1987scientific,
  title={Scientific discovery: Computational explorations of the creative processes},
  author={Langley, Pat},
  year={1987},
  publisher={MIT press}
}

@article{nianogo2015agent,
  title={Agent-based modeling of noncommunicable diseases: a systematic review},
  author={Nianogo, Roch A and Arah, Onyebuchi A},
  journal={American journal of public health},
  volume={105},
  number={3},
  pages={e20--e31},
  year={2015},
  publisher={American Public Health Association}
}

@article{jing2020agent,
  title={Agent-based simulation of autonomous vehicles: A systematic literature review},
  author={Jing, Peng and Hu, Hanbin and Zhan, Fengping and Chen, Yuexia and Shi, Yuji},
  journal={IEEE Access},
  volume={8},
  pages={79089--79103},
  year={2020},
  publisher={IEEE}
}

@article{willem2017lessons,
  title={Lessons from a decade of individual-based models for infectious disease transmission: a systematic review (2006-2015)},
  author={Willem, Lander and Verelst, Frederik and Bilcke, Joke and Hens, Niel and Beutels, Philippe},
  journal={BMC infectious diseases},
  volume={17},
  number={1},
  pages={612},
  year={2017},
  publisher={Springer}
}

@article{nugroho2023systematic,
  title={Systematic review of agent-based and system dynamics models for social-ecological system case studies},
  author={Nugroho, Supradianto and Uehara, Takuro},
  journal={Systems},
  volume={11},
  number={11},
  pages={530},
  year={2023},
  publisher={MDPI}
}

@inproceedings{belfrage2024simulating,
  title={Simulating change: A systematic literature review of agent-based models for policy-making},
  author={Belfrage, Michael and Lorig, Fabian and Davidsson, Paul},
  booktitle={2024 Annual Modeling and Simulation Conference (ANNSIM)},
  pages={1--13},
  year={2024},
  organization={IEEE}
}

@inproceedings{lame2019systematic,
  title={Systematic literature reviews: An introduction},
  author={Lame, Guillaume},
  booktitle={Proceedings of the design society: international conference on engineering design},
  volume={1},
  number={1},
  pages={1633--1642},
  year={2019},
  organization={Cambridge University Press}
}

@article{page2021prisma,
  title={The PRISMA 2020 statement: an updated guideline for reporting systematic reviews},
  author={Page, Matthew J and McKenzie, Joanne E and Bossuyt, Patrick M and Boutron, Isabelle and Hoffmann, Tammy C and Mulrow, Cynthia D and Shamseer, Larissa and Tetzlaff, Jennifer M and Akl, Elie A and Brennan, Sue E and others},
  journal={bmj},
  volume={372},
  year={2021},
  publisher={British Medical Journal Publishing Group}
}

@article{mitchener2025kosmos,
  title={Kosmos: An ai scientist for autonomous discovery},
  author={Mitchener, Ludovico and Yiu, Angela and Chang, Benjamin and Bourdenx, Mathieu and Nadolski, Tyler and Sulovari, Arvis and Landsness, Eric C and Barabasi, Daniel L and Narayanan, Siddharth and Evans, Nicky and others},
  journal={arXiv preprint arXiv:2511.02824},
  year={2025}
}

@misc{OpenAI2026,
  author = {{OpenAI}},
  title = {{OpenAI} {API}},
  url = {https://platform.openai.com/},
  year = {2026},
  note = {Version: GPT-4.1 and GPT-5.0, Accessed: 2026-01-23}
}

\section*{AUTHOR BIOGRAPHIES}

\noindent {\bf \MakeUppercase{Orhan Yagizer Cinar}} is an undergraduate student pursuing a B.S. degree in the Department of Computer Science at George Mason University. He works as a research assistant in the Department of Geography and Geoinformation Science. His research interests include large language models, computer vision, and AI-driven healthcare solutions, including computational models for disease detection and diagnosis. His email address is \email{ocinar@gmu.edu} and his website is \url{https://orhancinar.com}.  
\\

\noindent {\bf \MakeUppercase{Timur Emre Ozkose}} is an undergraduate student pursuing a B.S. degree in the Department of Computational and Data Sciences at George Mason University. He serves as a teaching assistant (STAR) in his department and as a research assistant in the Department of Geography and Geoinformation Science. His academic interests center on applying data analytics, data science, machine learning and AI tools, large
language models, and computational modeling to investigate real-world challenges, particularly those with a social dimension. His email address is \email{tozkose@gmu.edu}.
\\

\noindent {\bf \MakeUppercase{Emma Von Hoene}} is a Ph.D. candidate in the Department of Geography and Geoinformation Science at George Mason University in Fairfax, Virginia, USA. She holds a Master of Science in Geoinformatics and Geospatial Intelligence from GMU. Her research focuses on data-driven modeling and geospatial analysis. Her email address is \email{evonhoen@gmu.edu}.\\

\noindent {\bf \MakeUppercase{Amira Roess}} is a Professor of Epidemiology and Global Health at George Mason University's College of Public Health. Her research uses multi-disciplinary and multi-species field research, epidemiology and evaluation methods to study human, wildlife, pet and environmental determinants of health (One Health) in U.S. and global settings. Her email address is \email{aroess@gmu.edu} and her website is \url{https://publichealth.gmu.edu/profiles/aroess}.\\

\noindent {\bf \MakeUppercase{Taylor Anderson}} is an Associate Professor in the Department of Geography and Geoinformation Science at George Mason University. Her research focuses on modeling the spread of diseases in human and ecological systems. Her e-mail address is \email{tander6@gmu.edu} and her website is \url{https://science.gmu.edu/directory/taylor-anderson}.\\

\noindent {\bf \MakeUppercase{Hamdi Kavak}} is an Associate Professor in the Department of Computational and Data Sciences and Co-Director of Center for the Social Complexity at George Mason University. His research combines data science with modeling and simulation to investigate challenges in urban and social systems. His email address is \email{hkavak@gmu.edu} and his website is \url{https://hamdikavak.com/}.\\

\end{document}